\documentclass[journal]{IEEEtran}

\IEEEoverridecommandlockouts
\usepackage{cite}
\usepackage{amsmath,amssymb,amsfonts}
\usepackage{graphicx}
\usepackage{textcomp}
\usepackage{color}
\usepackage{xcolor}
\usepackage[ruled]{algorithm2e}
\usepackage{booktabs}
\usepackage{array}
\usepackage{algpseudocode}
\usepackage{amsmath}
\usepackage{tabularx}
\usepackage{makecell}
\newcolumntype{P}[1]{>{\centering\arraybackslash}p{#1}}
\newcolumntype{M}[1]{>{\centering\arraybackslash}m{#1}}
\usepackage{amssymb}
\usepackage{CJK}
\usepackage{multirow}
\usepackage{listings}
\usepackage{xcolor}
\usepackage{bm,graphicx,multirow,multicol,bm,bbm,psfrag,subfigure,color,mdframed,wasysym,subeqnarray}
\usepackage{hyphenat}
\usepackage{flushend}
\begin{document}
	
\title{
	A Novel Graph-based Motion Planner of Multi-Mobile Robot Systems with Formation and Obstacle Constraints
}

\author{Wenhang~Liu, Jiawei~Hu, Heng~Zhang, Michael~Yu~Wang,~\IEEEmembership{Fellow,~IEEE/ASME,} and Zhenhua~Xiong,~\IEEEmembership{Member,~IEEE}
	\thanks{
		This work was supported in part by the National Natural Science Foundation of China (U1813224), Ministry of Education China Mobile Research Fund Project (MCM20180703) and MoE Key Lab of Artificial Intelligence, AI Institute, Shanghai Jiao Tong University, China. \textit{(Corresponding author: Zhenhua Xiong.)}}
	\thanks{
		Wenhang~Liu, Jiawei~Hu, Heng~Zhang, and Zhenhua~Xiong are with the School of Mechanical Engineering, Shanghai Jiao Tong University, Shanghai, China (e-mail: liuwenhang@sjtu.edu.cn; hu\_jiawei@sjtu.edu.cn; zhanghengme\_sjtu@sjtu.edu.cn; mexiong@sjtu.edu.cn).
	}
	\thanks{
		Michael~Yu~Wang is with the Department of Mechanical and Aerospace Engineering, Monash University, Victoria 3800, Australia (e-mail: Michael.Y.Wang@monash.edu).
    }

}

\maketitle
	
\begin{abstract}

Multi-mobile robot systems show great advantages over one single robot in many applications.
However, the robots are required to form desired task-specified formations, making feasible motions decrease significantly.
Thus, it is challenging to determine whether the robots can pass through an obstructed environment under formation constraints, especially in an obstacle-rich environment.
Furthermore, is there an optimal path for the robots?
To deal with the two problems, a novel graph-based motion planner is proposed in this paper.
A mapping between workspace and configuration space of multi-mobile robot systems is first built, where valid configurations can be acquired to satisfy both formation constraints and collision avoidance.
Then, an undirected graph is generated by verifying connectivity between valid configurations.
The breadth-first search method is employed to answer the question of whether there is a feasible path on the graph.
Finally, an optimal path will be planned on the updated graph, considering the cost of path length and formation preference.
Simulation results show that the planner can be applied to get optimal motions of robots under formation constraints in obstacle-rich environments.
Additionally, different constraints are considered.

\end{abstract}


\section{Introduction}\label{sec1}

\IEEEPARstart {I}{n} recent years, multi-mobile robot systems (MMRS) have attracted increasing attention from robotics researchers.
Compared with a single robot, the robots can cooperate with each other to achieve better system robustness and flexibility.
Through collaboration among robots, complex tasks can be accomplished, such as search and rescue \cite{jennings1997cooperative, shree2021exploiting}, exploring \cite{nieto2014coordination}, assembling \cite{dogar2015multi} and transporting \cite{culbertson2021decentralized}.
Moreover, MMRS can replace the bigger single robot in scenarios with a restricted and cramped environment, where there is not enough space.
In many applications of MMRS mentioned above, the robots are usually required to form desired formations to achieve cooperation, such as the object will connect the robots to form a whole system during cooperative transporting.
Accordingly, the flexibility of each robot will be decreased since it has to satisfy formation constraints.
In the meanwhile, the robots also need to avoid obstacles when working, which requires them to be as flexible as possible.
Conflicts will arise if simultaneously considering formation constraints and obstacle avoidance when planning motions of robots, especially in a dense environment with multiple obstacles.
Therefore, it is a challenging problem to determine whether MMRS can pass through an obstructed environment under task-specified formation constraints.
Moreover, if it is possible, what is the optimal path for the robots to fulfill the task? 
Thus, we focus on multi-mobile robot motion planning problems under formation constraints in obstacle-rich environments in this paper.
A novel graph-based motion planner is proposed. A mapping between the workspace and configuration space of MMRS is first built, where valid configurations are acquired according to formation constraints and obstacle avoidance.
An undirected connected graph with boundary densification is then generated by verifying the connectivity between valid configurations.
Vertices and edges on the graph represent valid configurations and the connectivity between configurations, respectively.
Finally, the breadth-first search algorithm is adopted on the graph to quickly determine whether there is a feasible path, and the optimal path and motions of robots are computed considering path length and formation preference.
Cooperative object transportation is used as a typical example.
Simulation results in different obstacle-rich environments demonstrate the effectiveness and generality of the proposed planner.
Besides, the planner can be extended if there are additional constraints brought by the tasks.

The main contributions of this paper are threefold.
First, a novel mapping between workspace and configuration space of multi-mobile robot systems satisfying formation constraints is proposed.
Theoretical proofs show that the problems can be solved on the undirected connected graph generated by mapping.
Second, a boundary densification method is proposed on the connected graph to facilitate planning and improve the success rate for obstacle-rich environments.
Third, the planner can quickly determine whether it is feasible for MMRS to pass an obstacle-rich environment under formation constraints. Then, the optimal motions of the robots can be planned.

The rest of this paper is organized as follows.
In Section II, some related work is discussed.
In Section III, we define the basic concepts of multi-robot motion planning problems.
In Section IV, the proposed motion planner is explained in detail.
In Section V, the planner is employed for different formation constraints by a case study in object transportation.
The simulations and results are given in Section VI.
Section VII concludes the paper and outlines the future work.

\section{Related Work}\label{sec2}

In many studies of multi-robot motion planning, formation constraints among robots are not the key point.
For each robot, it regards other robots as dynamic obstacles and collaboration is not considered \cite{van2005roadmap, peasgood2008complete, kloder2006path}.
However, as tasks become more complex, the collaboration between multiple robots is inevitable, especially with formation constraints.

Generally, approaches in the previous works on motion planning of MMRS under formation constraints can be divided into two categories.
The first category is individually planning each robot while satisfying formation constraints and avoiding obstacles.
Under the combination of the two behaviors, the system can hold desired formations \cite{balch1998behavior}.
Based on the leader-follower and potential-based methods, desired formations with obstacle avoidance were achieved in \cite{seng2013distributed, chi2016strategy, wen2017formation}.
By combining the virtual structure and obstacle avoidance methods, \cite{rezaee2013decentralized} introduced an approach for multiple robots to maintain formations.
Similarly, the combination can be found in \cite{zhou2018agile}.
Besides, learning-based methods are also employed for the problems.
Formations could be adjusted independently by each robot to pass through obstacle areas based on reinforcement learning \cite{la2014multirobot, bai2021learning}.
In \cite{zhang2021task}, the task space, including formation constraints and obstacle avoidance, was decomposed into different convex regions through a global planner.
Then, configurations were planned through a local planner.
In the above research, the systems can all maintain desired formations tightly and move to the goal in environments without obstacles.
However, when it comes to obstacle avoidance, some uncontrollable deformation of system formations will appear more or less, which is not acceptable for some cooperative tasks.

The second category considers the multi-robot system as a whole 2D area, which is planned to be strictly separated from obstacles.
A system outlined rectangle approach was proposed in \cite{tang2018obstacle}, regarding the multi-robot system as a virtual rectangle to avoid obstacles.
Then, the problem was simplified to the motion planning of a single robot.
The idea can also be found in \cite{li2008real, jiao2015transportation}.
In \cite{alonso2015local} and \cite{alonso2017multi}, a constrained optimization method was presented for motion planning of multiple robots.
The robots could always form the desired formations in the largest convex obstacle-free region computed in the neighborhood of the system.
A region-based framework for multi-robot systems was introduced in \cite{roy2018multi}.
Based on virtual structure, robots moved and formed formations inside a changeable circular region always separated from obstacles.
In \cite{song2021herding}, formation control and change were achieved by caging behaviors.
The whole system was planned by the rapidly-exploring random tree for obstacle avoidance.
Sometimes, the multi-robot systems have to be considered as a whole due to special tasks, as in \cite{fink2008multi, wang2016force, machado2016multi, koung2021cooperative}.
However, for the above research, it is over-constrained since robots can actually be separated.
At least, the system could have crossed obstacles instead of merely bypassing obstacles as a whole.
Therefore, the system loses part of its obstacle avoidance ability.

It is noted that current research did not explicitly consider the formation constraints. Thus, the deformation of systems during motion cannot be limited to a certain range of tasks.
The conflict between forming formations and avoiding obstacles has not been fully handled.
In some situations where rigid or less flexible connections among the robots are required, these methods are unsuitable.
2D region-based methods are safe for obstacle avoidance. However, it is only suitable for a few obstacles environment.
Therefore, it is worth finding a method for multi-robot motion planning in obstacle-rich environments, at the same time keeping formation constraints.

In multi-robot motion planning studies, graph theory is a powerful tool \cite{solovey2014k,yu2015intractability,adler2015efficient}.
In \cite{peasgood2008complete}, based on a graph and spanning tree representation, a multi-phase approach was described.
Through the relationship between multi-robot planning problems and multi-flow problems, integer linear programming models were proposed to compute optimal paths for multiple robots on connected graphs \cite{yu2013planning}.
Their models were further extended in \cite{yu2016optimal} in connection with different minimization objectives in path planning.
The optimal solution or approximate optimal solution could be quickly calculated on the connected graphs.
In \cite{salvado2021combining}, goal allocation and motion planning were achieved on roadmaps for non-holonomic robots. 

In the above studies with graph theory, the robots are more likely to be planned individually, where the confliction between waypoints is the only connection among multiple robots \cite{wu2020multi,motes2020multi,le2019multi,han2018sear}.
However, formation constraints are not considered.
Therefore, in order to solve the problem through graph theory, the mapping between configuration space and workspace needs to be studied \cite{wan2016efficient}.
Moreover, the configuration space of MMRS should be further investigated, especially in the case of integrated obstacle avoidance and formation constraints.

\section{Problem Formulation}\label{sec3}

In this work, we consider multi-mobile robot motion planning under both obstacle-rich environments and formation constraints.
The mobile robots are assumed to be holonomic, and they move on a 2D plane.
The typical task is to transport an object from the initial position to goal position.

\begin{table}[h]
	\centering
	\caption{Definition of Some Important Notations}
	\label{table1}
	\begin{tabular}{  p{1cm}<{\centering} | p{6.5cm} }
		\toprule[1.5pt]
		$r_i$ & The position of $i$th mobile robot. \\ [2pt]
		$p$ & The center of multi-mobile robot systems. \\ [2pt]
		$\mathcal P$ & The plane where mobile robots are located. \\[2pt]
		$\mathcal S$ & Formation constraints of multi-robot systems. \\ [2pt]
		$\theta$ & The angle of formations. \\ [2pt]
		$\mathcal C$ & Whole configuration space of a system. \\ [2pt]
		$\bm c$ & Configurations of a system,  $\bm c = \left\{  x,y,\theta,\mathcal S \right\}$. \\ [2pt]
		$\bm c^{s}$ & Configurations of a system in the same formation shape,  $\bm c^{s} = \left\{  x,y,\theta \right\}$. \\ [2pt]
		$o_i$ & The position of $i$th obstacle. \\ [2pt]
		$\mathcal W(o_i)$ & Space occupied by obstacle $o_i$. \\ [2pt]
		$\mathcal W(r_i)$ & Space occupied by robot $r_i$. \\ [2pt]
		$\mathcal C_{free}$ & A set of valid configurations, $\mathcal C_{free} \subset \mathcal C$. \\ [2pt]
		$\mathcal C_{free}^{s}$ & A set of valid configurations in the same formation shape.  \\ [2pt]
		$\partial \mathcal C_{free}^{s}$ & The boundary of $ \mathcal C_{free}^{s} $. \\ [2pt]
		$\mathcal{N}(\bm c^{s}, \xi )$ & Neighborhood configurations of $\bm c^{s}$. \\ [2pt]
		$\bm{\Omega}( \bm c^{s,p} )$ & Valid angles of the same formation $\mathcal S$ with the same $p$. \\ [2pt]
		\bottomrule[1.5pt]
	\end{tabular}
\end{table}

\begin{figure}
	\includegraphics[width=\columnwidth]{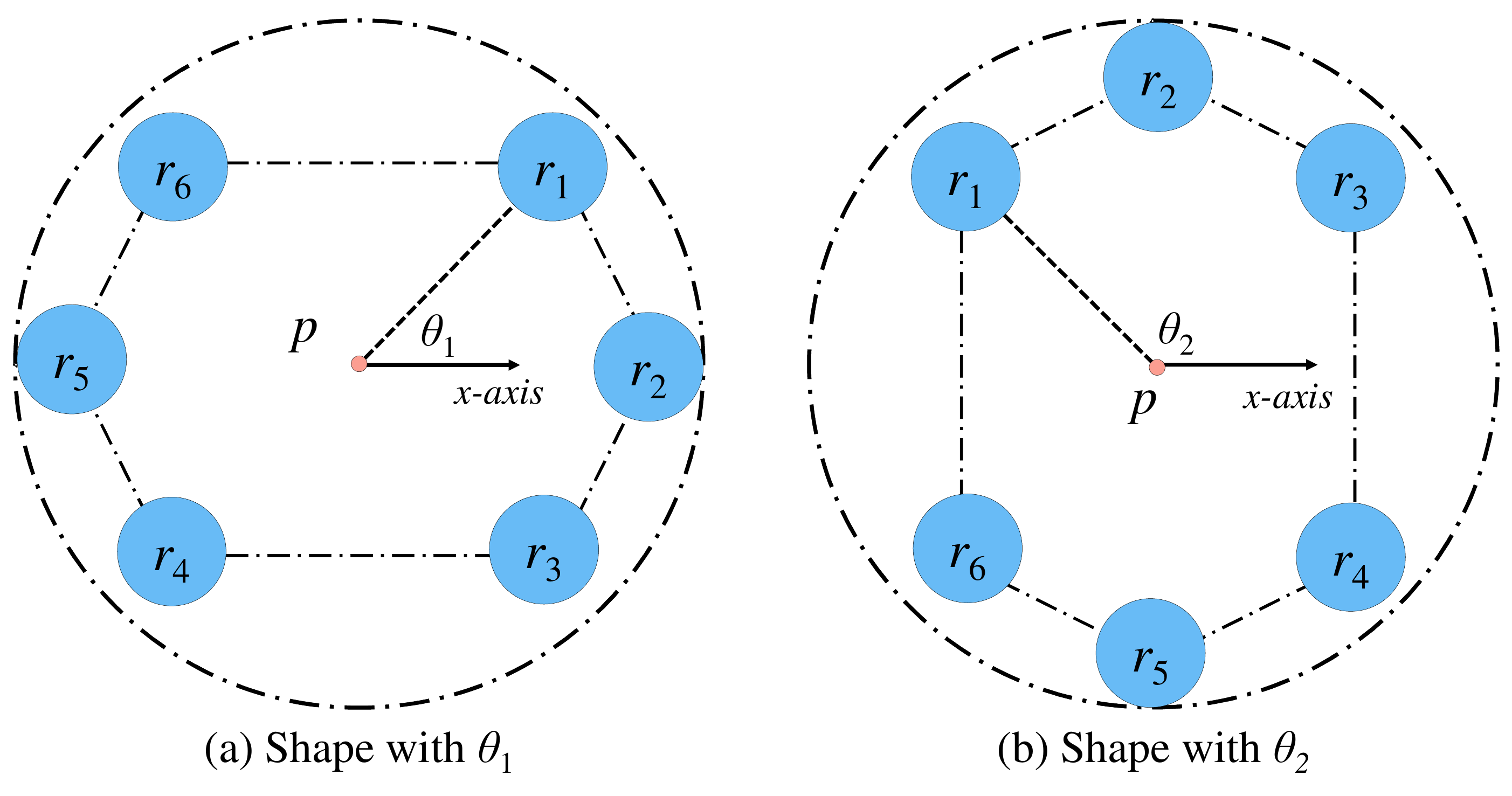}
	\centering
	\caption{
		The same formation shape with different angles. $p$ is the same in (a) and (b), while positions of robots are different.
	}
	\label{fig_formation}       
\end{figure}

Some important notations are listed in Table I for convenience. In a workplace $ \mathcal{W}\subset{{\mathbb{R}}^{3}} $, the position of the $i$th mobile robot is denoted as ${r}_{i}\in{{\mathbb{R}}^{2}}, i=1,\ldots ,n$, and $n$ is used to denote the number of mobile robots.
The plane where robots are located is denoted by $\mathcal{P} \subset {{\mathbb{R}}^{2}}$.
The center of multi-robot systems is denoted as $ p = [x,y] := \sum\limits_{i=1}^{n}\frac{r_i}{n} \in {{\mathbb{R}}^{2}} $, which is used to describe the absolute location of systems in the workplace.
Formation $s \in \mathcal{S}$ formed by robots is defined as follows.
\begin{equation}
	\label{eq1}
	\begin{aligned}  
		 s := & \left\{d_{ij}^* \right\}  \quad i,j = 1,\dots,n \quad i \ne j  \\
		 d_{ij}^* = & {\left\| r_i - r_j \right\|}_2   \\		 
	\end{aligned}
\end{equation}
where $s$ is the desired formation shape of multi-robot systems, which consists of relative positions between robots.
Since $p$ is not changed by the rotation of the same formation shape, multiple positions of robots may exist.
As shown in Fig.~\ref{fig_formation}, it can be seen that the two formation shapes are the same, but $r_i$ is different from each other.
Therefore, $\theta \in \mathbb{R}$ is used to denote the shape angle and distinguish positions of robots.
To sum up, the configuration $\bm c$ of multi-mobile robot systems can be described as $\bm c =  \left\{p,\theta,\mathcal{S} \right\}$, and the whole configuration space is denoted as $\mathcal C$.
Once $\bm c$ is given, the position of each robot in the workspace can be determined as follows.
\begin{equation}
	\label{eq2}
	\left\{  {r}_{1},\ldots , {r}_{n}    \right\} = \mathcal{S}^{-1} ( p ,\theta )
\end{equation}
where $\mathcal{S}^{-1}$ indicates positions of robots are calculated by the relative position relationship within the formation shape.

Now, let ${o}_{i}\subset{{\mathbb{R}}^{2}}, i=1,\ldots ,m$ denotes the position of the $i$th obstacle, and $m$ is the number of obstacles.
$\mathcal W(o_i) \subset {{\mathbb{R}}^{3}} $ is used to denoted the workspace occupied by the $i$th obstacle.
Similarly, $\mathcal W(r_i)$ is the workspace occupied by the $i$th robot. 
The set of valid configurations $\mathcal{C}_{free}$ is defined as follows.
\begin{align}
	\label{eq3}
	\mathcal{C}_{free} :=& \left\{  \bm c \ |  \  \mathcal R \cap \mathcal O = \varnothing \right\}   \notag \\
	{\rm where} \ \mathcal{R}  =&  \mathcal W(r_i) \cup \dots \cup \mathcal W(r_n)        \\
	\mathcal{O}  =&  \mathcal W(o_i) \cup \dots \cup \mathcal W(o_m)     \notag
\end{align}
In other words, $\bm c \in \mathcal{C}_{free}$ indicates that the desired formation exists, satisfying that all robots are not in contact with obstacles.
Planning configurations in $\mathcal{C}_{free}$ ensures safety during movement.
If we directly plan the motion of each robot with formation constraints and obstacle avoidance, it will suffer from the curse of dimensionality.
Generally, it is as high as ${\mathbb{R}}^{2n}$.
To avoid this, configurations of systems will be planned first, after which paths of robots can be quickly computed.
It can be found that configurations of a system consist of absolute information $\left\{ x,y,\theta\right\}$ and relative information $\mathcal S$.
Therefore, planning configurations of a system is divided into two parts.
The first part is to plan $\mathcal S$, which is more dependent on formation constraints and will be further discussed.
$\mathcal S$ is strongly related to applications and tasks.
The second part is to plan $\left\{ x,y,\theta\right\}$ for driving the system on $\mathcal P$, which is more concerned in this section.

For a multi-mobile system, the planning problem considered in this paper can be formulated as follows.
\begin{equation}
	\label{eq4}
	\begin{aligned} 
		\text{Input}:& \  \mathcal O, \mathcal S \\
		\text{Output}:& \left\{   \bm c_1,\dots,\bm c_k    \right\} \\
		\text{Subject to} & \begin{cases}
			\bm c_i \in \mathcal C_{free}, \  \bm c_k \ \text{is the goal} \\\\
			\bm c_i \ \text{and} \ \bm c_{i+1} \ \text{are connected}
		\end{cases}
	\end{aligned}
\end{equation}

Since it is extremely hard to obtain a real continuous space of $\mathcal{C}_{free}$ in the graph, we will solve the problem \eqref{eq4} by translating it using the discretization method. 
If two configurations can be switched to each other through all available configurations in $\mathcal{C}_{free}$, they are defined as connected.
Therefore, the first issue to be studied is the connectivity between configurations after discretizing the workspace.

\begin{figure}
	\includegraphics[width=\columnwidth]{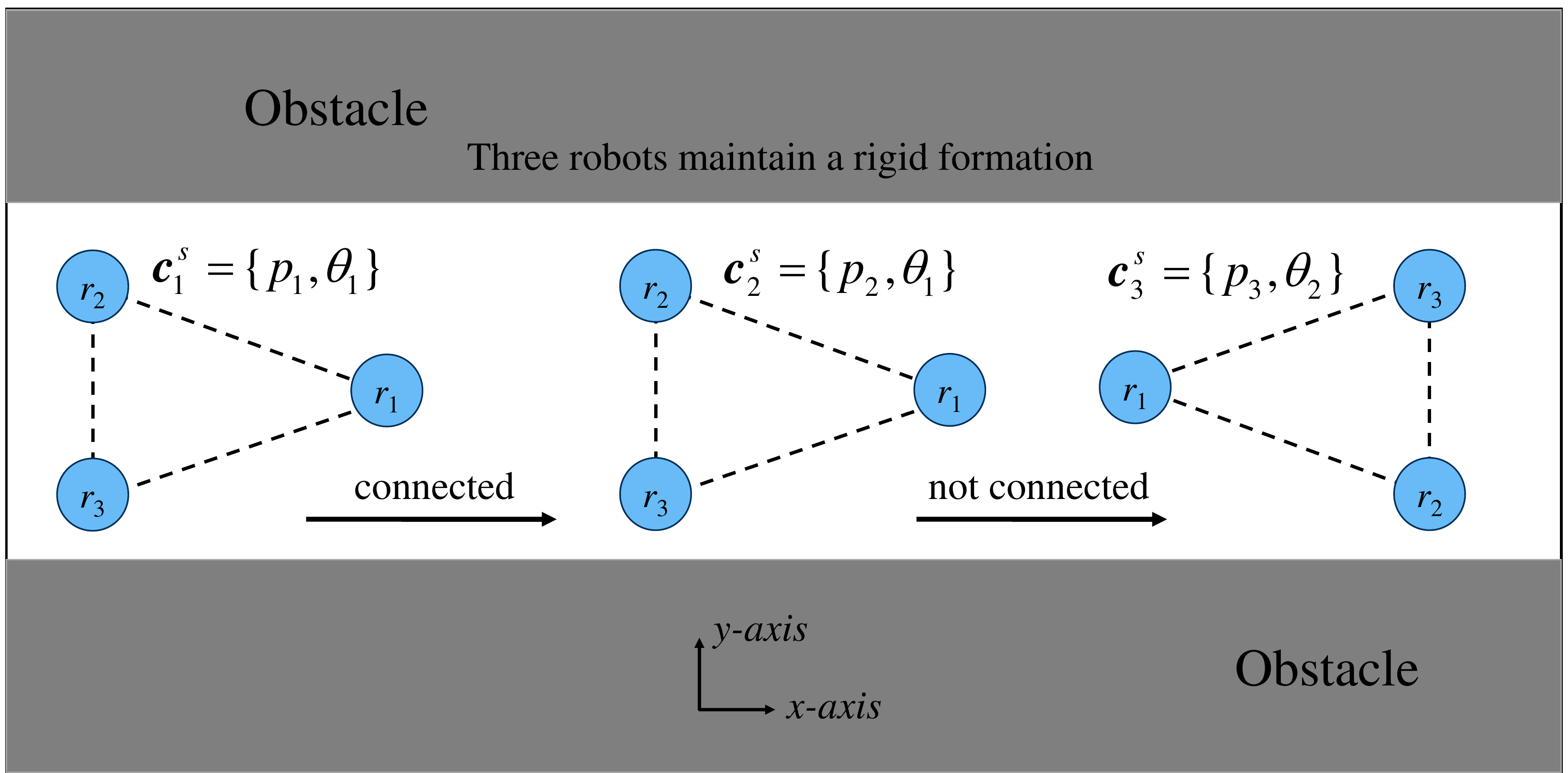}
	\centering
	\caption{
		Different valid configurations in $\mathcal{C}^{s}_{free}$ of a system composed of three mobile robots.
	}
	\label{fig_configurations}       
\end{figure}

First, given a formation shape $s$, the configuration space of the same formation is denoted as $\bm c^{s} = \left\{  x,y,\theta   \right\}$.
The set of valid configurations of this formation is defined as $\mathcal{C}^{s}_{free}$.
As shown in Fig.~\ref{fig_configurations}, $\bm c_1^{s}$ and $ \bm c_2^{s}$ are connected because the system can move from $\bm c_1^{s}$ to $\bm c_2^{s}$ while keeping the formation continuously unchanged.
However, $ \bm c_2^{s}$ and $ \bm c_3^{s}$ are not connected although they are both in $\mathcal{C}^{s}_{free}$.
They can switch to each other only if ignoring the formation constraint.
Hence, it is necessary to study the connectivity between different configurations before planning.

\textbf{Theorem 1}. \textit{For any $\bm c^{s} \in \mathcal{C}^{s}_{free}$, $\bm c^{s}$ is connected in a neighborhood area 
$\mathcal{N}(\bm c^{s}, \xi ):= \left\{   \tilde{\bm c}^{s}  \right\}$, configurations in neighborhood area are defined as follows.}
\begin{equation}
	\label{eq5}
	\begin{aligned}  
			{\left\| (\tilde{x} - {x})^2 + (\tilde{y} - {y}^2 \right\|}_2 &< \xi \\
			{\rm abs}(\tilde{\theta} - \theta ) &< \xi ,\ \xi>0 
 	\end{aligned}
\end{equation}

\textit{Proof}: $\forall \bm c^{s} = \left\{x,y,\theta \right\} \in \mathcal{C}^{s}_{free}$, from the definition \eqref{eq3}, all robots are separated from obstacles although their distances may be very small.
In this case, a specific obstacle-free space $\mathcal T_i$ can be found for the $i$th robot, which satisfies
\begin{equation}
	\label{eq6}
	\mathcal W(r_i) \subset \mathcal{T}_i, \mathcal{T}_i \cap \mathcal{O} = \varnothing
\end{equation}
$\mathcal T_i$ is a bounded open set.
Since \eqref{eq6} is available for each robot, the whole formation can move or rotate freely in any direction if $\mathcal W(\tilde{r}_i)$ still stays in $\mathcal T_i$ after movement, which will generate new valid configurations.
Therefore, $ \bm c^{s}$ is connected with these configurations, and the area is denoted as $\mathcal{N}(\bm c^{s}, \xi )$.
Obviously, $\xi$ is strongly related to $\mathcal T_i$ and is relatively bigger in an environment without obstacles.
It should be pointed out that definition \eqref{eq5} is a little conservative because the whole formation may be able to move bigger than $\xi$ in some directions.
However, the conservative definition does not affect overall connectivity.
$\hfill\blacksquare$

Theoretically, the real continuous space of $\mathcal{C}_{free}$ can be obtained with infinite discretization.
However, the calculation cost is unacceptable.
In order to get closer to the real $\mathcal{C}_{free}$ and improve the success rate of subsequent planning, a boundary densification method is introduced.
Naturally, we need to look at how to find the boundary $\partial \mathcal{C}^{s}_{free}$.

Now, we use  $\bm{\Omega}( \bm c^{s,p} )$ to denote all valid angles of the same formation with the same $p$.
The maximum range of $\bm{\Omega}( \bm c^{s,p} )$ is $ [0,2\pi) $.
A weak assumption is made for the following statement.
When the formation rotates in a fixed $p$, robots and obstacles cannot critically contact all the time.
This assumption is quite weak because it occurs only when the shapes of robots and obstacles are always inscribed.

\textbf{Theorem 2}. \textit{$\bm c^{s}$ approaches the boundary $\partial \mathcal{C}^{s}_{free}$ if and only if $\bm{\Omega}( \bm c^{s,p} )$ converges to a specific angle. In other words, $\bm{\Omega}( \bm c^{s,p} ) = \theta^{*}$ when $\bm c^{s,p} \in \partial \mathcal{C}^{s}_{free}$.}

\begin{figure}
	\includegraphics[width=\columnwidth]{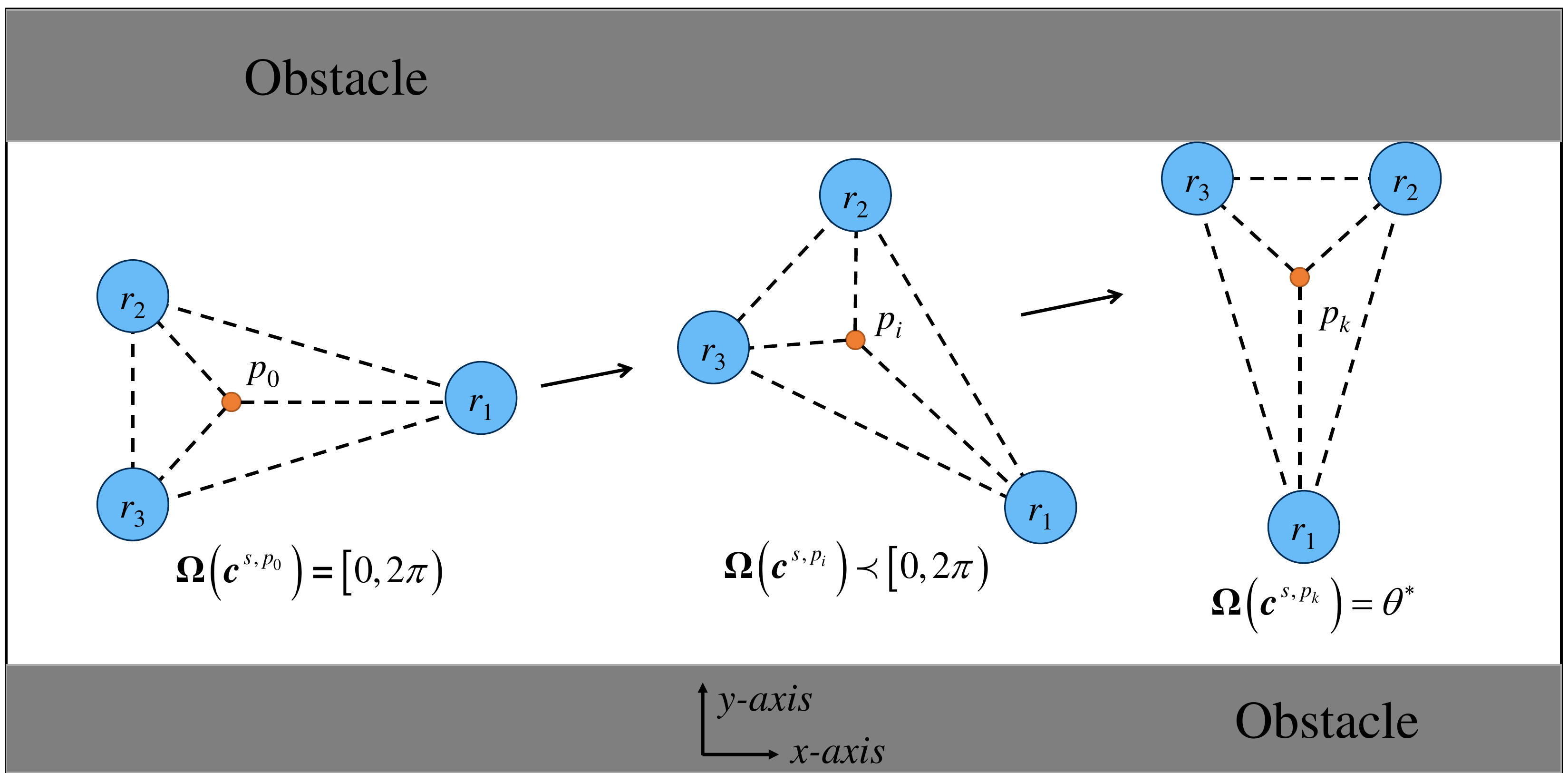}
	\centering
	\caption{
		The change of $\bm{\Omega}$ when approaching the boundary.
	}
	\label{fig_boundary}       
\end{figure}

\textit{Proof}: Denote the initial position $p = p_{0}$ and $\bm c^{s,p_0} \in \mathcal{C}^{s}_{free}$.
According to Theorem 1, $\bm c^{s,p_0}$ is connected in $\mathcal N( \bm c^{s,p_0}, \xi_{0} )$, $\bm c^{s,p_1}$ is randomly chosen in $\mathcal N( \bm c^{s,p_0}, \xi_{0} )$,  then do
\begin{equation}
	\label{eq7}
    ( p_{k+1} - p_{k} ) = \lambda_k ( p_{1} - p_{0} )   \quad  \bm c^{s,p_{k+1}} \in \mathcal N( \bm c^{s,p_k}, \xi_{k} )
\end{equation}
where $\lambda_k$ is a positive coefficient, and this moves the system in one direction.
During \eqref{eq7}, $\mathcal N( \bm c^{s,p_k}, \xi_{k} )$ may become bigger or smaller depending on the environment.
$\bm c^{s,p_{k+1}}$ approaches the boundary $\partial \mathcal{C}^{s}_{free}$ if and only if $\xi_{k+1} \to 0$.
That means, $\mathcal{T}_{i} \to \mathcal W(r_i)$ at least for one robot and the robot is fairly close to obstacles.
Due to the assumption, shapes of robots and obstacles are not inscribed.
Therefore, the formation cannot rotate around $p_{k+1}$ and $\xi_{k+1} \to 0$.
As a result, $\bm{\Omega}( \bm c^{s,p_{k+1}} )$ will approach the intermittent value $\theta^{*}$, which indicates that $\bm{\Omega}( \bm c^{s,p} ) = \theta^{*}$ when $\bm c^{s,p} \in \partial \mathcal{C}^{s}_{free}$.
A case is shown in Fig.~\ref{fig_boundary}, the system forms a rigid formation, where $\bm{\Omega}(\bm c^{s,p_0}) = [0,2\pi )$ and $\bm{\Omega}(\bm c^{s,p_k}) \to \theta^{*}$.
It can be seen that $\bm{\Omega}$ gradually reduces to a specific angle when approaching the boundary.
$\hfill\blacksquare$

\begin{figure}
	\includegraphics[width=\columnwidth]{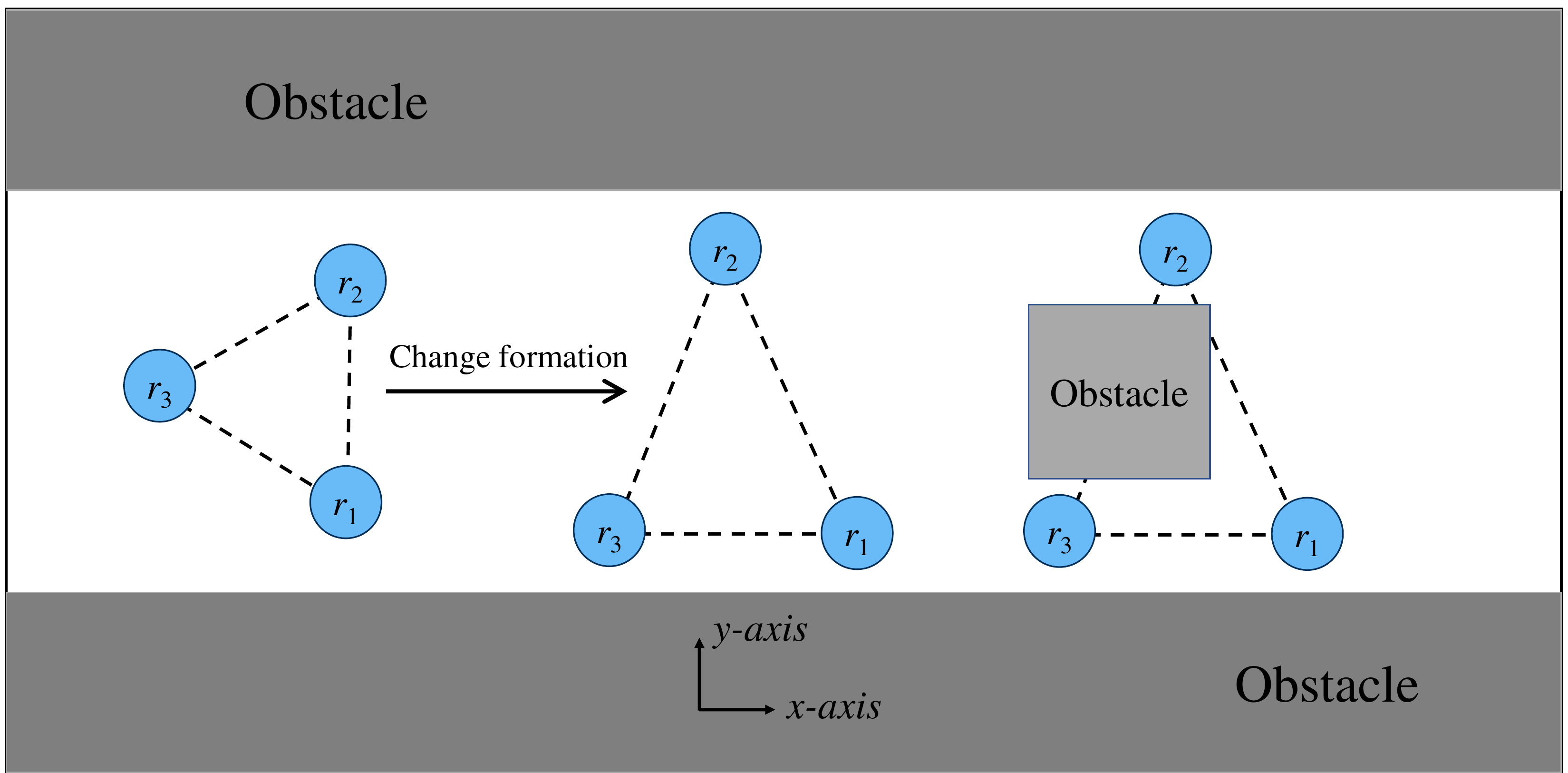}
	\centering
	\caption{
		The system is forced to change formation in order to pass obstacles.
	}
	\label{fig_changeformation}       
\end{figure}

Theorem 1-2 are enrolled for unchanged formations.
The multi-robot system can move from the initial position to goal position with the desired formation if two configurations are both in $\mathcal{C}^{s}_{free}$ and connected.
But in some situations, the formation is expected or required to change.
As shown in Fig.~\ref{fig_changeformation}, the system can never pass the obstacle if the formation remains unchanged.
Therefore, the switch between different formations also needs to be studied.
For the sake of brevity but without loss of generality, we assume two formations and the difference between them is small enough (This statement is not rigorous, we will explain further in the following).

\begin{figure*}[t]
	\includegraphics[width=\textwidth]{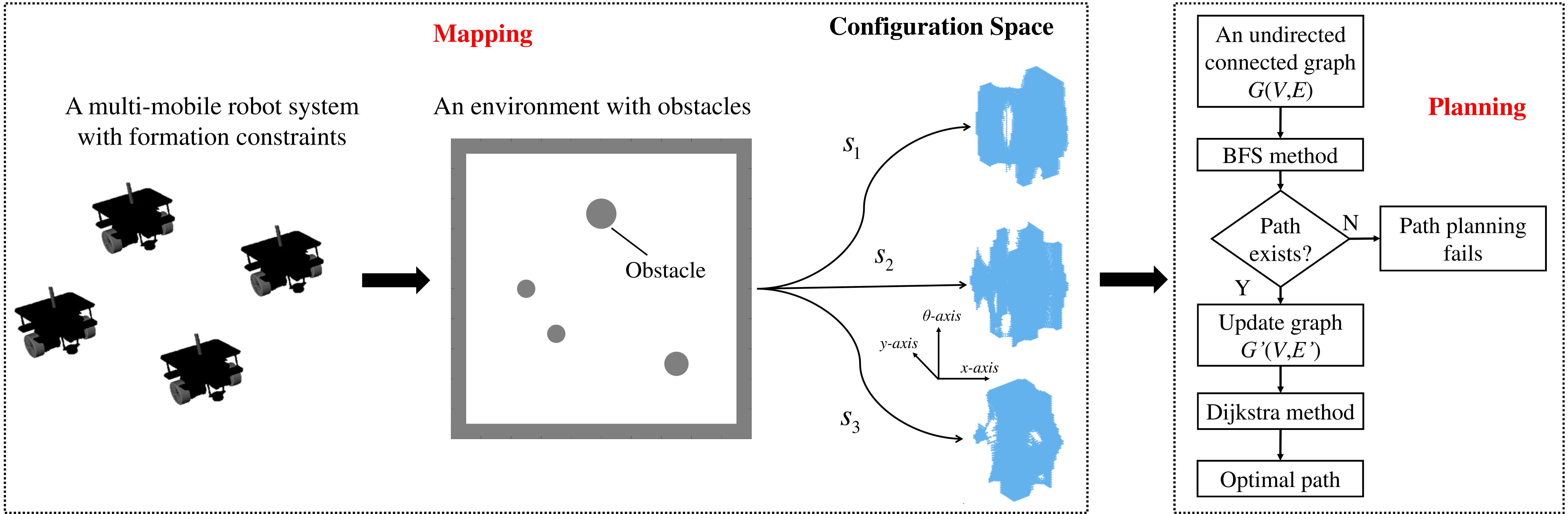}
	\centering
	\caption{
		The general framework of our planner. The workspace is mapped into the valid configuration space with boundary densification. An undirected connected graph is generated by verifying the connectivity of valid configurations. Whether a path exists is quickly determined on the graph without cost. The optimal path is planned on the updated graph considering path length and formation preference. Then, motions of robots are computed on the path.
	}
	\label{fig_planner}       
\end{figure*}

\textbf{Theorem 3}. \textit{If $\bm c^{s_1,p} \in \mathcal{C}^{s_1}_{free}$, $\bm c^{s_2,p} \in \mathcal{C}^{s_2}_{free}$, and $\exists \theta_{0} \in \bm{\Omega}(\bm c^{s_1,p})  \cap  \bm{\Omega}(\bm c^{s_2,p}) $, two formations are switchable in $p$.}

\textit{Proof}: Assume robots form the first formation $s_1$ with $\theta_{0}$ in $p$.
The position of $i$th robot is $r^{s_1}_1$.
According to Theorem 1, $\mathcal W(r^{s_1}_i) \subset \mathcal{T}_{i}$.
The difference between two formations is small enough means $\mathcal W(r^{s_1}_i) \subset \mathcal{T}_{i}$ and $\mathcal W(r^{s_2}_i) \subset \mathcal{T}_{i}$ for all robots.
Then, the formation can switch from $s_1$ to $s_2$.
Though Theorem 3 is for two formations with relatively small differences, it can be used to test whether two formations are switchable.
This is more practical in later applications.
$\hfill\blacksquare$

In the above statement, we adequately describe configurations of multi-mobile robot systems.
The connectivity between configurations of the same formation and switching between different formations have been studied in Theorem 1-3.
Based on these, our planner is designed, which will be further introduced in the next section.

\section{The Proposed Planner}\label{sec4}

For a given multi-mobile robot system and an environment with obstacles, we assume that the desired formations $\mathcal S$ have been determined.
The framework of our planner is shown in Fig.~\ref{fig_planner}.
The environment is firstly discretized into cells, which are mapped into the valid configuration space of the system.
Areas of the configuration space near the boundary are further refined.
Then, the connectivity of two configurations is verified to generate an undirected connected graph.
Initially, there is no specific cost in the graph, and the breadth-first search method \cite{rose1976algorithmic} can be applied to quickly decide whether a feasible path exists or not.
If the answer is yes, a cost function based on the path lengths of robots and formation preference is assigned to update the graph.
Finally, the Dijkstra method \cite{jonker1987shortest} is applied to find the optimal path in the updated graph.
Our planner is able to be combined with different formation constraints in obstacle-rich environments.
Moreover, this planner can be extended in some special situations with additional constraints, which will be discussed in the next section.

\subsection{Mapping from Workspace to Configuration Space}

The first step of our planner is to map the points $p$ on the plane of robots to the configurations $\bm c$ of the multi-mobile robot system.
Mapping between workspace and configuration space has been widely used in robot planning.
In this paper, full mapping is considered to find the optimal planning path.
The mapping $\mathcal{M}$ between $\mathcal P$ and $\mathcal C$ is built as follows.
\begin{equation}
	\label{eq8}
	\begin{aligned} 
		\mathcal{M}(p):=&\left\{ \bm c \ | \ x_{\bm c} = x_p, y_{\bm c} = y_p, \bm c \in \mathcal C \right\} \\
		\mathcal{M}(\bm c):=&\left\{ p \ | \ x_p = x_{\bm c}, y_p = y_{\bm c}, p \in \mathcal P \right\}
	\end{aligned}
\end{equation}
In our multi-mobile robot systems, configurations are denoted as $\bm c = \left\{ x,y,\theta,\mathcal S\right\}$.
Obviously, $\mathcal{M}(p)$ is a one-to-many mapping, and $\mathcal{M}(\bm c)$ is a many-to-one mapping.

To solve the problem through graph theory, we need to discretize the space.
As explained before, $x,y \in \mathcal P$ and $\theta \in \bm \Omega$ are used to denote the absolute locations of robots.
$\mathcal P$ is discretized by $g$, and $\bm \Omega \subset [0,2\pi)$ is discretized by $\alpha$.
Discrete scales $g$ and $\alpha$ are crucial parameters in the planner, determining the distance between adjacent robot waypoints.
In some cases, discrete scale is also called resolution or grid size.
$g$ is chosen according to the following rules.

\noindent (1) $g \in [g_{\text {min}}, g_{\text {max}}]$.

\noindent (2) $g_{\text {min}}$ is related to computational cost and accuracy.
The control accuracy of robots is limited and can never reach such a resolution as small as possible.
Therefore, $g_{\text {min}}$ cannot be selected too small.
Otherwise, it will be impractical or computationally unacceptable. 

\noindent (3) $g_{\text {max}}$ is related to the connectivity of configurations.
This is decided by the size of obstacles and robots.
Path planning may fail if $g_{\text {max}}$ is selected too big.

\begin{algorithm}[t]
	\caption{Configurations Mapping}
	\label{alg1}
	\LinesNumbered
	\KwIn{Multi-robot systems; Discrete scales $g$, $\alpha$, $g_\text{min}$;
		Formation constraints $\mathcal{S}$;
		Environment with obstacles $\left\{ o_1,\dots,o_m \right\}$.}
	\KwOut{Valid configurations $\bm c \in \mathcal C_{free}$.}
	Initializing: $\mathcal C_{free} \gets \varnothing$; \\
	$p_i \gets \texttt{PlaneDiscretized}(g,\alpha)$;\\
	\ForEach{$p_i$}
	{
		$\bm c \gets \mathcal{M}(p_i)$; \\
		\If{ \rm{\texttt{ValidConfiguration}$(\bm c)$} is true }
		{  $\bm c$ is added to $\mathcal C_{free}$; \\
		}
	}
	\ForEach{$\bm c \in \mathcal C_{free}$}
	{
		\If{ \rm {\texttt{BoundaryDetect}}$(\bm c)$ is true}
		{  $p_i \gets \mathcal{M}(\bm c)$; \\
			$p_{\text {new}} \gets \texttt{BoundaryDiscretized}(p_i,g_\text{min},\alpha)$; \\
			\ForEach{\rm $p_{\text {new}}$}
			{ Repeat 4-7;
			}
		}
	}
\end{algorithm}

\begin{figure}[t]
	\includegraphics[width=\columnwidth]{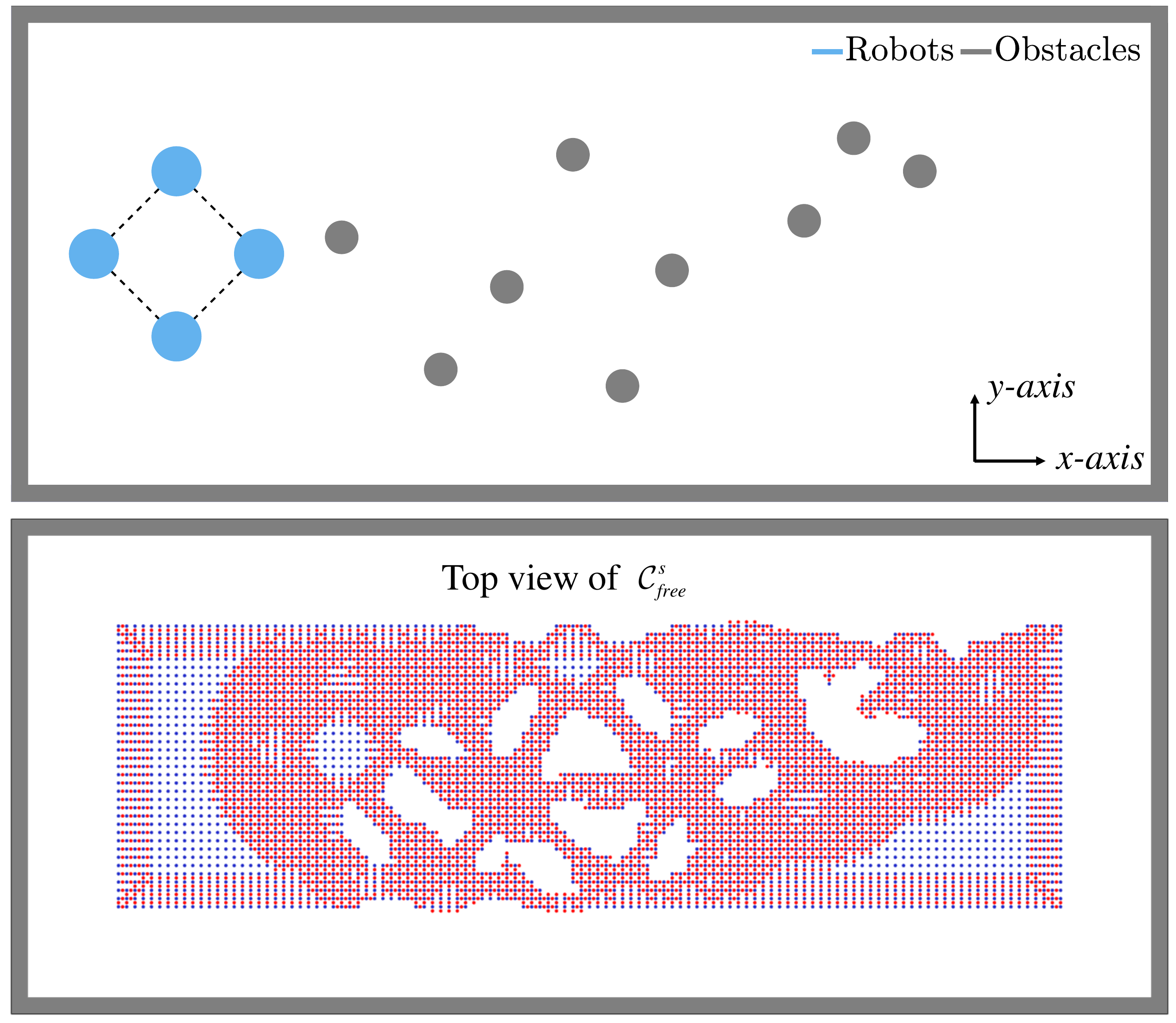}
	\centering
	\caption{
		A plane with boundary densification. It is firstly discretized by $g$, generating the blue points, which can be mapped into valid configurations. After finding areas near the boundary, it is further discretized by $g_\text{min}$, generating the red points.
	}
	\label{fig_discretescale}       
\end{figure}

In general, $g$ is set within a reasonable range in order to solve the problem practically and practicably.
The choice of $\alpha$ is similar to $g$.
It cannot be too small because of the accuracy, nor can it be too big to cross obstacles during rotation.
Once discrete scales are determined, for each formation constraint $s$, the 3D configuration space $\mathcal C^{s}_{free}$ can be built.
Valid configurations near the boundary $\partial \mathcal C^{s}_{free}$ are selected.
Then, $p$ near the boundary in $\mathcal P$ is computed by $\mathcal{M}(\bm c)$.
For areas on $\mathcal P$ near the boundary, the minimal discrete scale $g_\text{min}$ will be used to discretize them further.
The newly resulting $p$ is added to the mapping.
In this way, the most realistic situation of $\mathcal C^{s}_{free}$ can be approximated to the greatest extent without excessively increasing the computation.
An example is given in Fig.~\ref{fig_discretescale}.
The blue points represent $p$ discretized by $g$, and these points can be mapped into valid configurations.
After finding the $\mathcal{M}(p)$ near $\partial \mathcal C^{s}_{free}$, the plane is further discretized by $g_\text{min}$, generating the red points.
Twice discretization can make the boundary denser, thereby increasing the success rate of planning, which will be demonstrated by simulations later. 
The whole process of mapping is shown in Algorithm \ref{alg1}.

\subsection{Connected Graph Generator}

The second step is to generate an undirected connected graph through valid configurations.
Graph theory is usually used to describe a relationship between certain things.
Let $\mathcal{G=(V,E)}$ denotes a connected graph, where $\mathcal{V} = \left\{ v_{i} \right\}$ is the vertex set and $\mathcal{E} = \left\{ e_{ij}:=( v_{i},v_{j}) \right\}$ is the edge set.
Graphs can be divided into directed graphs and undirected graphs, depending on $e_{ij}$ and $e_{ji}$.
If $e_{ij} = e_{ji}$ for all edges, the graph is undirected.
Otherwise, the graph is directed.
In path planning, the points or positions of robots are stored in $\mathcal{V}$, and costs between points are stored in $\mathcal{E}$.
The task is to schedule a path from the initial vertex to the goal vertex.
Due to previous research, many quick and convenient planning methods can be directly applied.

In our problem, $v_i$ represents a configuration, which has been built by Algorithm \ref{alg1}.
Then, it needs to find $e_{ij}$, which represents the cost of configuration switch.
To improve planning efficiency, whether two configurations can be transformed into each other will be determined primarily.
The specific cost is not considered at first.
The connectivity between configurations can be verified according to Theorem 1 and 3.
We exclusively consider the connectivity of two adjacent configurations, which means only one dimension is different in $\bm c$.
Although this is incomplete in finding $e_{ij}$, it does not affect the overall connected graph and will significantly decrease the computation.
$e_{ij}$ is added to the edge set $\mathcal{E}$ if $v_{i}$ and $v_{j}$ are connected.
Now, an undirected connected graph $\mathcal{G}$ is obtained.
In $\mathcal{V}$, the initial and goal configurations are denoted as $v_\text{init}$ and $v_\text{goal}$, respectively.
For $\mathcal{G}$, we can use the Breadth First Search (BFS) method to quickly search whether there is a path from $v_\text{init}$ to $v_\text{goal}$.
If it exists, we will calculate the specific cost in $\mathcal{E}$ and then find the optimal path.
If it does not exist, path planning fails, and all subsequent calculations will be saved.

There are three options for configuration change: movement, rotation, and formation change.
A cost function is designed based on the consideration of path lengths of robots and formation preference as follows.
\begin{equation}
	\label{eq9}
	e_{ij} = \begin{cases}
		\sum_{k=1}^{N} \lambda_{s} \left\|  {r}_{k}^{ v_{i}} -  {r}_{k}^{v_{j}}   \right\|_2    \ \text{ move or rotate } \\\\
		\sum_{k=1}^{N}  \left\|  {r}_{k}^{v_{i}} -  {r}_{k}^{v_{j}}   \right\|_2     \   \text{change formation}
	\end{cases}
\end{equation}
where $\sum_{k=1}^{N} \left\| {r}_{k}^{v_{i}} - {r}_{k}^{v_{j}}   \right\|_2 $ is the total distance travelled by all robots.
$\lambda_{s}$ is the indicator of preference.
The system will try to maintain the desired formation with a smaller $\lambda_{s}$.
As a result, we can change the tendency of formations by controlling the relative relationship between $\lambda_{s}$ of different formations.
After calculating the specific cost, $\mathcal{G}$ is updated to $\mathcal{G}'$.
Dijkstra algorithm can be enrolled to find the optimal path in $\mathcal{G}'$.
The whole planning process is shown in Algorithm \ref{alg2}.
Searching twice by different methods is trying to save time and computational cost.

\begin{algorithm}[t]
	\caption{Planning on Graph}
	\label{alg2}
	\LinesNumbered
	\KwIn{Valid configurations $\bm c \in \mathcal C_{free}$; Cost function.}
	\KwOut{Robot path $R_1,\dots,R_n$.}
	Initializing: $R_1,\dots,R_n \gets \varnothing$, $\mathcal{G(V,E)} \gets \varnothing$. \\
	\ForEach{$\bm c_i \in \mathcal C_{free}$}
	{   $\bm c_i$ is added to $\mathcal{V}$ as $v_i$; \\
	}
	\For{$v_i, v_j \in \mathcal{V}$}
	{
		\If{ \rm \texttt{ConnectionDetect}$(v_i,v_j)$ is true}
		{  $e_{ij}$ is added to $\mathcal{E}$; \\
		}
	}
	$\Gamma \gets$ \texttt{BFS}($\mathcal{G}$); \\
	\eIf{ $\Gamma$ exists }
	{  $\mathcal{G}' \gets \texttt{CostFunction}(\mathcal{G})$; \\
		$\Gamma \gets$ \texttt{Dijkstra}($\mathcal{G}'$); \\
		$R_1,\dots,R_n \gets \texttt{RobotPath}(\Gamma)$;\\
		\Return $R_1,\dots,R_n$;
	}
	{   \Return $R_1,\dots,R_n$;
	}
\end{algorithm}

\section{Case Study in Object Transportation}\label{sec5}

In this paper, we take multi-robot object transportation as a typical example.
There are not only constraints brought by formations but also by objects in transportation \cite{tuci2018cooperative}.
Therefore, it is more challenging than other scenarios.
Generally, multi-robot transportation systems are composed of mobile robots and the carrier.
Carriers can be mobile robot itself \cite{chen2015occlusion, yufka2015formation, ebel2020design}, deformable sheet \cite{hunte2019collaborative, hu2022multi, hunte2021collaborative}, or manipulator \cite{petitti2016decentralized, ren2020fully, wu2016collaboration}, which result in different system model and formation constraints.
Still and all, they can be solved by our planner, and the first two will be used as examples in this paper.

Now, a little modification is made to the previous definition.
$p = [x,y]$ is used to denote the position of object, and $\mathcal W(p)$ is used to denote the space occupied by object and carriers.
There is a certain relationship as follows.
\begin{equation}
	\label{eq10}
	p = \mathfrak{F}( {r}_{1},\ldots ,{r}_{n}) 
\end{equation}
where $\mathfrak{F}$ is the mapping between positions of robots and the object, which is directly determined by the carrier.
As explained before, formation constraints $\mathcal S$ are the internal relative information of multi-robot systems.
Therefore, if we care about the relative positional relationship within the system, Eq.\eqref{eq10} can be rewritten as follows.
\begin{equation}
	\label{eq11}
	\hat{p} = \mathfrak{F}( \mathcal S ) 
\end{equation}
where $\hat{p}$ is the relative position of objects inside the system.
Generally, Eq.\eqref{eq10} is used to describe the absolute locations of the transportation system in the workplace, and Eq.\eqref{eq11} is used for relative positions within the system.
Since there are new constraints brought by the object and carriers, the definition of valid configurations \eqref{eq3} is changed as follows.
\begin{equation}
	\label{eq12}
	\mathcal C_{free} := \left\{  \bm c \ |  \  \mathcal W(p) \cap \mathcal R \cap \mathcal O = \varnothing \right\}  
\end{equation}
The newly added constraints ensure that the object does not collide with robots and obstacles during transportation.
Desired formations are determined according to Eq.\eqref{eq10}-\eqref{eq12}.
For example, two cases are given later.

\subsection{Transportation without Carrier}

In some systems, there are no special carriers and mobile robots themselves are regarded as carriers.
These systems are cheap and easy to control.
This is the simplest case because robots are often physically attached to objects, which means the formation constraint $\mathcal S$ is fixed.
During transportation, relative positions between robots and objects are invariably kept as the initial.
Consequently, valid configuration space is reduce to $\mathcal C^{\mathcal S}_{free} \subset {{\mathbb{R}}^{3}}$.
In the fixed $\mathcal S$, the whole system can be considered as a rigid body to avoid obstacles.

\subsection{Transportation with Deformable Sheet}

\begin{figure}[t]
	\includegraphics[width=\columnwidth]{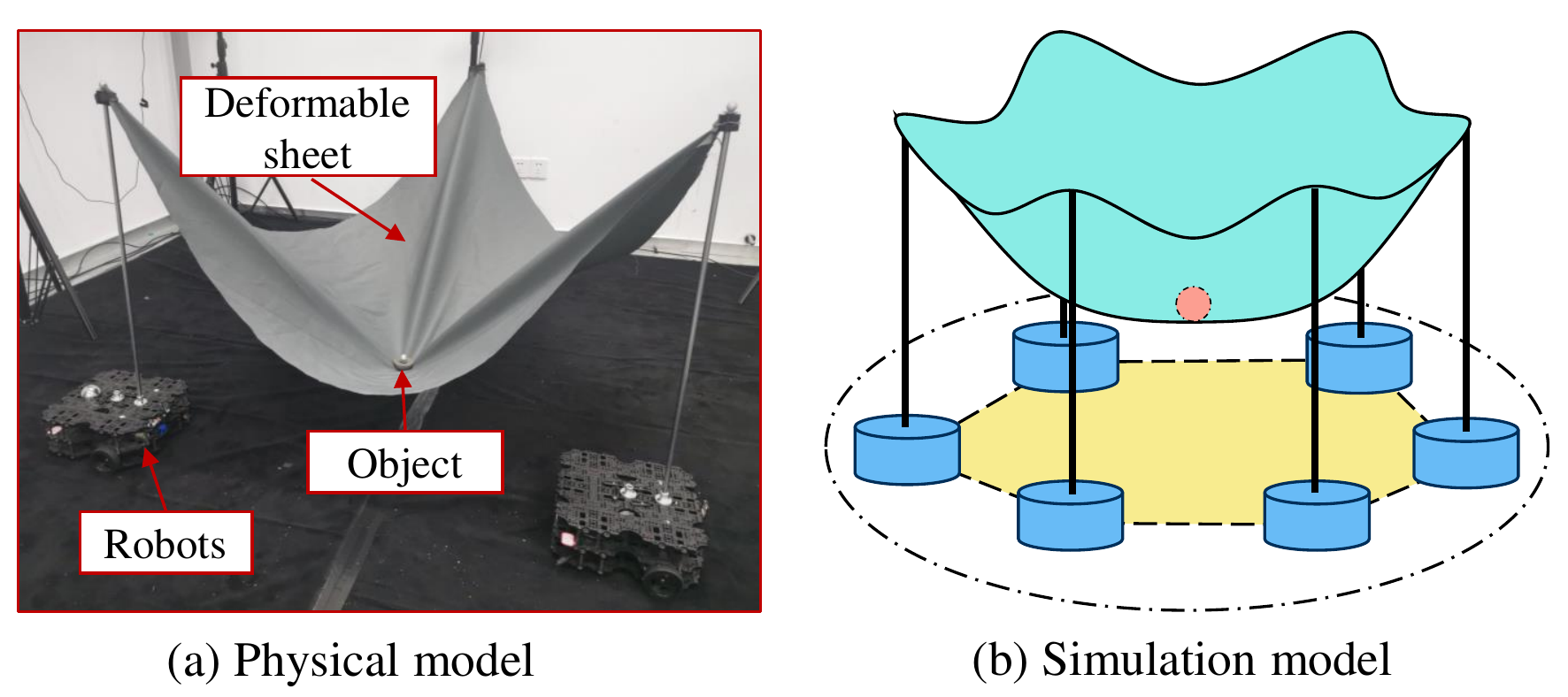}
	\centering
	\caption{
		Multi-robot transportation with a deformable sheet as the carrier.
	}
	\label{fig_deformablesheet}       
\end{figure}

\begin{figure*}[t]
	\includegraphics[width=\textwidth]{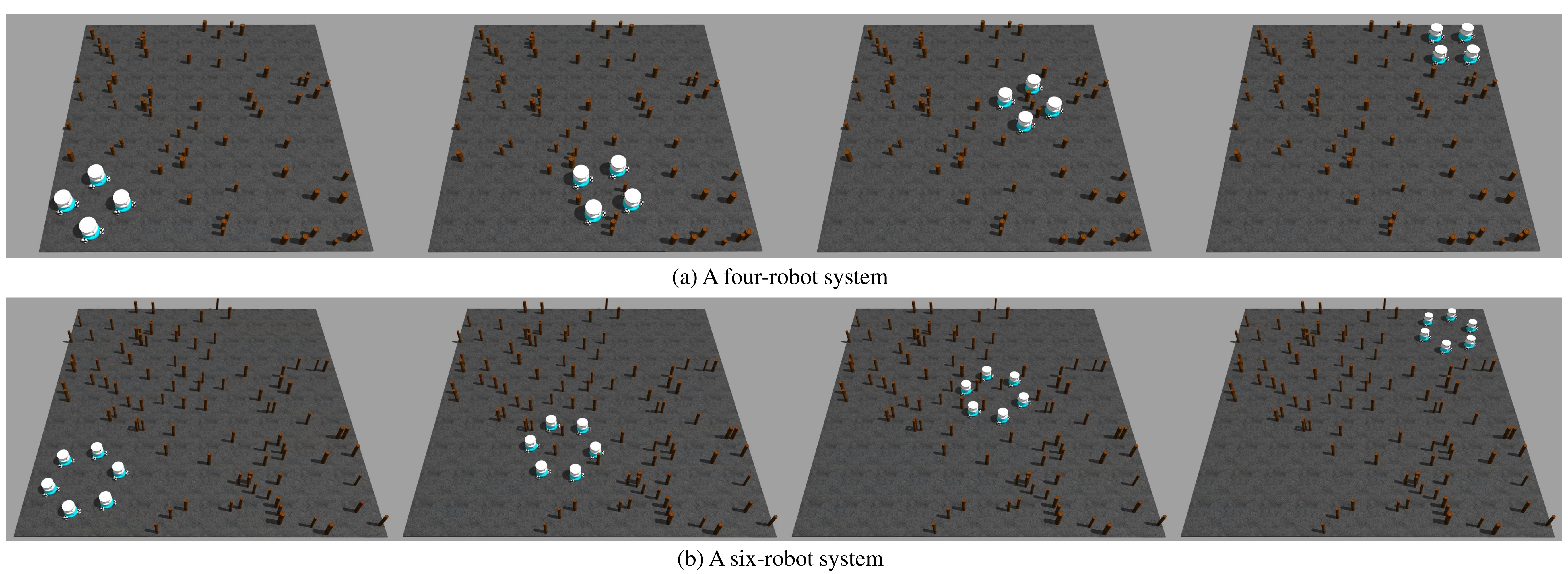}
	\centering
	\caption{
		Multi-mobile robot systems traverse obstacle-rich environments while maintaining rigid formations.
		(a) Four robots form a square in a 10m$\times$10m environment. There are 65 obstacles with random positions and radii.
		(b) Six robots form a hexagon in a 15m$\times$15m environment. There are 90 obstacles with random positions and radii.
	}
	\label{fig_rigid}       
\end{figure*}

\begin{figure*}[t]
	\includegraphics[width=\textwidth]{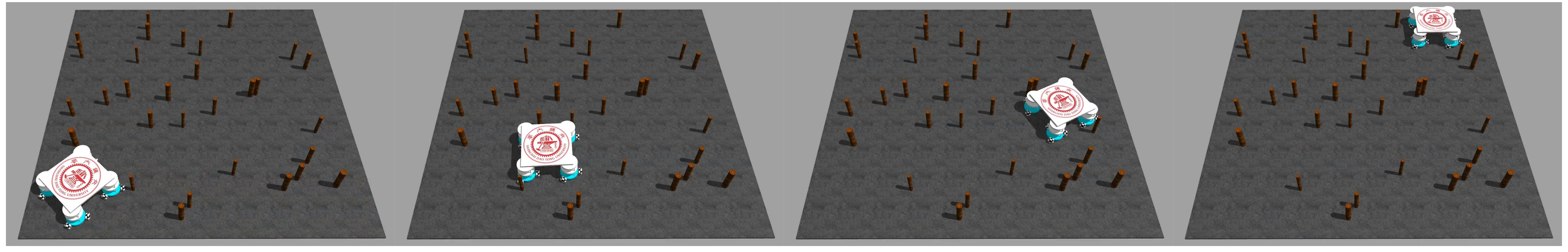}
	\centering
	\caption{
		A multi-mobile robot system composed of four robots transports a square object in a 10m$\times$10m environment.
		There are 30 obstacles with random positions and radii. The system is forced to bypass obstacles like a 2D whole.
	}
	\label{fig_without}       
\end{figure*}

Multi-robot transportation with a deformable sheet as the carrier is shown in Fig.~\ref{fig_deformablesheet}.
The transported objects are placed on the deformable sheet held by robots.
These systems are universal to different shapes of objects and the position of object can be manipulated by adjusting relative positions between robots.

The model of these systems has been studied in our previous work \cite{hu2022multi}.
A computational approach based on the Virtual Variable Cables Model(VVCM) is presented for calculating Eq.\eqref{eq10}, which simplifies the system to a robots-cables-payload system.
Then, $\mathfrak{F}$ is replaced by VVCM as follows.
\begin{equation}
	\label{eq13}
	p = \textit{VVCM}({r}_{1},\ldots ,{r}_{n})
\end{equation}
Eq.\eqref{eq13} assumes that the position of object is at the state of minimum gravitational potential energy solved by convex optimization.
It should be pointed out that the minimal system is composed of three robots.
For the minimal systems, Eq.\eqref{eq10} can be directly calculated by VVCM.
For systems with a higher number of robots, there are multiple local static equilibrium states, resulting in multiple solutions.
To deal with the multiple solutions, an optimal formation function is introduced.
\begin{equation}
	\begin{aligned}
		\label{eq14}
		J(s) &= J_\text{trans} + J_\text{pass} + J_\text{cross}  \\
		s^* &= \text{argmin} \ J(s)
	\end{aligned}
\end{equation}
where $J_\text{trans}$ ensures the safety and stability of systems during transportation.
Valid configurations are included in $J_\text{pass}$ and $J_\text{cross}$, where the former is to bypass obstacles, the latter is to cross obstacles.
For an environment with obstacles, optimal formations are obtained by Eq. \eqref{eq13} and \eqref{eq14}.
With the generated optimal formations $\mathcal S$, the planning problem can be solved by our planner.

\section{Simulations and Discussions}\label{sec6}

\subsection{Simulation Setup}

In this section, we have designed three scenarios for multi-mobile robot systems under different formation constraints to test our planner.
These systems are placed in an environment containing a large number of randomly generated obstacles.
All our simulations are performed on a computer with Intel Core i7-10700 CPU at 2.90 GHz (16 cores) and 64GB RAM.
The results are obtained by Gazebo and Rviz.

\subsection{Simulation Results}

\textit{1) Maintaining Rigid Formations}

In the first part, mobile robots are required to form a rigid formation while traversing an environment with obstacles.
Since the formation is rigid, which indicates that $\mathcal S$ is given and onefold, the valid configuration of systems is reduced to three-dimension.
We set up an environment of 10m$\times$10m.
The initial position of formation center is (1.5,1.5)m, and the target is (8.5,8.5)m.
On the premise of ensuring that the initial and target position exists, some obstacles with random positions are generated in the environment, whose radiuses are also randomly selected between 0.05m and 0.1m.
The radius of each robot is 0.35m.
The results are shown in Fig.~\ref{fig_rigid}.
In Fig.~\ref{fig_rigid}(a), four robots try to maintain a square formation.
There are 65 obstacles in the environment.
The average path length of each robot is 19.1599m on the optimally planned path.
Similarly, in Fig.~\ref{fig_rigid}(b), six robots try to maintain a hexagon, and there are 90 obstacles in an environment of 15m$\times$15m.
The initial position of formation center is (2.5,2.5)m while the target is (12.5,12.5)m.
The average path length of each robot is 24.7397m.
It can be seen that the two multi-mobile robot systems are able to traverse the obstacle-rich environments while holding desired formations invariably unchanged.
To our best knowledge, no existing planner for multi-robot systems can deal with these environments.

\begin{table*}[t]
	\centering
	\fontsize{6.5}{8}\selectfont
	\caption{Repeat {\rm Fig.~\ref{fig_rigid}(a)} under Different Conditions without Boundary Densification}
	\label{table2}
	\begin{tabular}{m{0.8cm}<{\centering}|m{1.2cm}<{\centering}m{0.9cm}<{\centering}m{0.9cm}<{\centering}m{0.9cm}<{\centering}|m{1.2cm}<{\centering}m{0.9cm}<{\centering}m{0.9cm}<{\centering}m{0.9cm}<{\centering}|m{1.2cm}<{\centering}m{0.9cm}<{\centering}m{0.9cm}<{\centering}m{0.9cm}<{\centering}}
		\toprule[1.5pt]
		\multirow{3}{*}{Obstacles}&
		\multicolumn{4}{c|}{Discrete Scale: 0.08m 0.04rad}&
		\multicolumn{4}{c|}{Discrete Scale: 0.05m 0.04rad}&
		\multicolumn{4}{c}{Discrete Scale: 0.025m 0.04rad} \\
		&\multirow{2}{*}{Success Rate}&\multicolumn{3}{c|}{Average Time (s)}&\multirow{2}{*}{Success Rate}&\multicolumn{3}{c|}{Average Time (s)}&\multirow{2}{*}{Success Rate}&\multicolumn{3}{c}{Average Time (s)}\\
		&&Mapping&Planning&Total&&Mapping&Planning&Total&&Mapping&Planning&\multicolumn{1}{c}{Total} \\ [2pt]
		\midrule[1.5pt]
		10    & 100/100 & 2.12479  & 0.50741  & 2.63220  & 100/100 & 5.26545  & 1.39201  & 6.65746  & 100/100 & 20.87622  & 5.98285  & 26.85907  \\
		20    & 99/100 & 2.18363  & 0.34327  & 2.52690  & 99/100 & 5.42517  & 0.82054  & 6.24571  & 99/100 & 21.50460  & 3.87619  & 25.38079  \\
		30    & 93/100 & 2.22332  & 0.22335  & 2.44667  & \textcolor{blue}{93/100} & 5.53143  & 0.59595  & \textcolor{blue}{6.12738}  & \textbf{94/100} & 21.93061  & 2.76713  & \textbf{24.69774}  \\
		40    & \textcolor{red}{67/100} & 2.25200  & 0.14657  &\textcolor{red}{2.39857}  & \textcolor{blue}{70/100} & 5.59970  & 0.40005  & \textcolor{blue}{5.99975}  & \textbf{77/100} & 22.30289  & 1.80580  & \textbf{24.10869}  \\
		50    & \textcolor{red}{36/100} & 2.28206  & 0.10722  & \textcolor{red}{2.38928}  & \textcolor{blue}{41/100} & 5.67768  & 0.31582  & \textcolor{blue}{5.99350}  & \textbf{50/100} & 22.64238  & 1.31906  & \textbf{23.96144}  \\
		60    & 23/100 & 2.31072  & 0.07745  & 2.38817  & \textcolor{blue}{23/100} & 5.78103  & 0.21294  & \textcolor{blue}{5.99397}  & \textbf{26/100} & 22.95013  & 0.91771  & \textbf{23.86784}  \\
		70    & 2/100 & 2.32667  & 0.05235  & 2.37902  & \textcolor{blue}{2/100} & 5.84270  & 0.15498  & \textcolor{blue}{5.99768}  & \textbf{5/100} & 23.17927  & 0.66619  & \textbf{23.84546}  \\
		80    & 0/100 & 2.34902  & 0.04201  & 2.39103  & 1/100 & 5.89965  & 0.11418  & 6.01383  & 1/100 & 23.34135  & 0.50529  & 23.84664  \\
		90    & 1/100 & 2.36974  & 0.03097  & 2.40071  & 1/100 & 5.95067  & 0.10586  & 6.05653  & 1/100 & 23.69105  & 0.42129  & 24.11234  \\
		100   & 0/100 & 2.38406  & 0.02386  & 2.40792  & 0/100 & 5.98362  & 0.07432  & 6.05794  & 0/100 & 23.90271  & 0.31573  & 24.21844  \\
		\bottomrule[1.5pt]
	\end{tabular}
\end{table*}

\begin{table*}[t]
	\centering
	\fontsize{6.5}{8}\selectfont
	\caption{Repeat {\rm Fig.~\ref{fig_rigid}(a)} under Different Conditions with Boundary Densification}
	\label{table3}
	\begin{tabular}{m{0.8cm}<{\centering}|m{1.2cm}<{\centering}m{0.9cm}<{\centering}m{0.9cm}<{\centering}m{0.9cm}<{\centering}|m{1.2cm}<{\centering}m{0.9cm}<{\centering}m{0.9cm}<{\centering}m{0.9cm}<{\centering}|m{1.2cm}<{\centering}m{0.9cm}<{\centering}m{0.9cm}<{\centering}m{0.9cm}<{\centering}}
		\toprule[1.5pt]
		\multirow{3}{*}{Obstacles}&
		\multicolumn{4}{c|}{Discrete Scale: 0.08m/0.025m 0.04rad}&
		\multicolumn{4}{c|}{Discrete Scale: 0.05m/0.025m 0.04rad}&
		\multicolumn{4}{c}{Discrete Scale: 0.025m 0.04rad} \\
		&\multirow{2}{*}{Success Rate}&\multicolumn{3}{c|}{Average Time (s)}&\multirow{2}{*}{Success Rate}&\multicolumn{3}{c|}{Average Time (s)}&\multirow{2}{*}{Success Rate}&\multicolumn{3}{c}{Average Time (s)}\\
		&&Mapping&Planning&Total&&Mapping&Planning&Total&&Mapping&Planning&\multicolumn{1}{c}{Total} \\ [2pt]
		\midrule[1.5pt]
		10    & 100/100 & 2.73023  & 1.19019  & 3.92042  & 100/100 & 6.57826  & 2.66315  & 9.24141  & 100/100 & 20.87622  & 5.98285  & 26.85907  \\
		20    & 99/100 & 2.77770  & 0.91127  & 3.68897  & 99/100 & 6.50419  & 1.97885  & 8.48304  & 99/100 & 21.50460  & 3.87619  & 25.38079  \\
		30    & 93/100 & 2.80989  & 0.74911  & 3.55900  & \textcolor{blue}{94/100} & 6.42345  & 1.46426  & \textcolor{blue}{7.88771}  & \textbf{94/100} & 21.93061  & 2.76713  & \textbf{24.69774}  \\
		40    & \textcolor{red}{69/100} & 2.76320  & 0.53806  & \textcolor{red}{3.30126}  & \textcolor{blue}{72/100}& 6.38639  & 1.06474  & \textcolor{blue}{7.45113}  & \textbf{77/100} & 22.30289  & 1.80580  & \textbf{24.10869}  \\
		50    & \textcolor{red}{37/100} & 2.75822  & 0.40120  & \textcolor{red}{3.15942} & \textcolor{blue}{47/100} & 6.48182  & 0.84711  & \textcolor{blue}{7.32893}  & \textbf{50/100} & 22.64238  & 1.31906  & \textbf{23.96144}  \\
		60    & 23/100 & 2.67300  & 0.29691  & 2.96991  & \textcolor{blue}{24/100} & 6.47801  & 0.64271  & \textcolor{blue}{7.12072}  & \textbf{26/100} & 22.95013  & 0.91771  & \textbf{23.86784}  \\
		70    & 2/100 & 2.65702  & 0.22995  & 2.88697  & \textcolor{blue}{3/100 }& 6.33030  & 0.45257  & \textcolor{blue}{6.78287}  & \textbf{5/100} & 23.17927  & 0.66619  & \textbf{23.84546}  \\
		80    & 0/100 & 2.62448  & 0.17087  & 2.79535  & 1/100 & 6.28102  & 0.34993  & 6.63095  & 1/100 & 23.34135  & 0.50529  & 23.84664  \\
		90    & 1/100 & 2.58503  & 0.12706  & 2.71209  & 1/100 & 6.33555  & 0.29529  & 6.63084  & 1/100 & 23.69105  & 0.42129  & 24.11234  \\
		100   & 0/100 & 2.50583  & 0.07612  & 2.58195  & 0/100 & 6.25361  & 0.22073  & 6.47434  & 0/100 & 23.90271  & 0.31573  & 24.21844  \\
		\bottomrule[1.5pt]
	\end{tabular}
\end{table*}

It should be pointed out that our planner may also fail in these environments since all obstacles are randomly generated.
Therefore, we repeat lots of times for the system composed of four robots under different numbers of obstacles.
Moreover, in order to study the effect of discrete scales in the planner, we also repeat under different scales.
The results are given in Table \ref{table2}.
For each time, obstacles are randomly generated while the initial and goal positions are fixed, and the environment is discretized only once by given scales.
It can be seen that under the same discrete scales, as the number of obstacles increases, the success rate generally decreases, and the time spent on mapping from workspace to configuration space increases while the time spent on planning on the connected graph decreases.
Essentially, the whole configuration space is limited and the same for determined discrete scales.
When obstacles increase, the time for validating valid configurations increases, and obviously, the size of valid configuration space decreases.
Then, the size of connected graph decreases, resulting in less time for planning.
It can also be seen that for the same number of obstacles, the success rate and total time are both larger under smaller discrete scales.
However, the success rate cannot be significantly improved by minifying discrete scales when the number of obstacles is more than 80.
This means the workplace is too harsh for the system.

As shown in Table \ref{table2}, decreasing the discrete scales will significantly increase the computational cost.
Therefore, we try our proposed boundary densification method.
The minimal scale $g_\text{min} = 0.025$, and the results are given in Table \ref{table3}.
The relationship between the time spent on mapping and the number of obstacles is no longer linear since the time on \texttt{BoundaryDetect} in Algorithm \ref{alg1} is irregular.
Compared with Table \ref{table2}, for the same environments, the success rate is higher with boundary densification, as indicated by the numbers marked in red and blue.
Sometimes the success rate is even close to the case with minimal scales.
However, the computational cost is far less than discretization directly by minimal scales.
Therefore, the proposed boundary densification method can facilitate planning.

\begin{figure*}[t]
	\includegraphics[width=\textwidth]{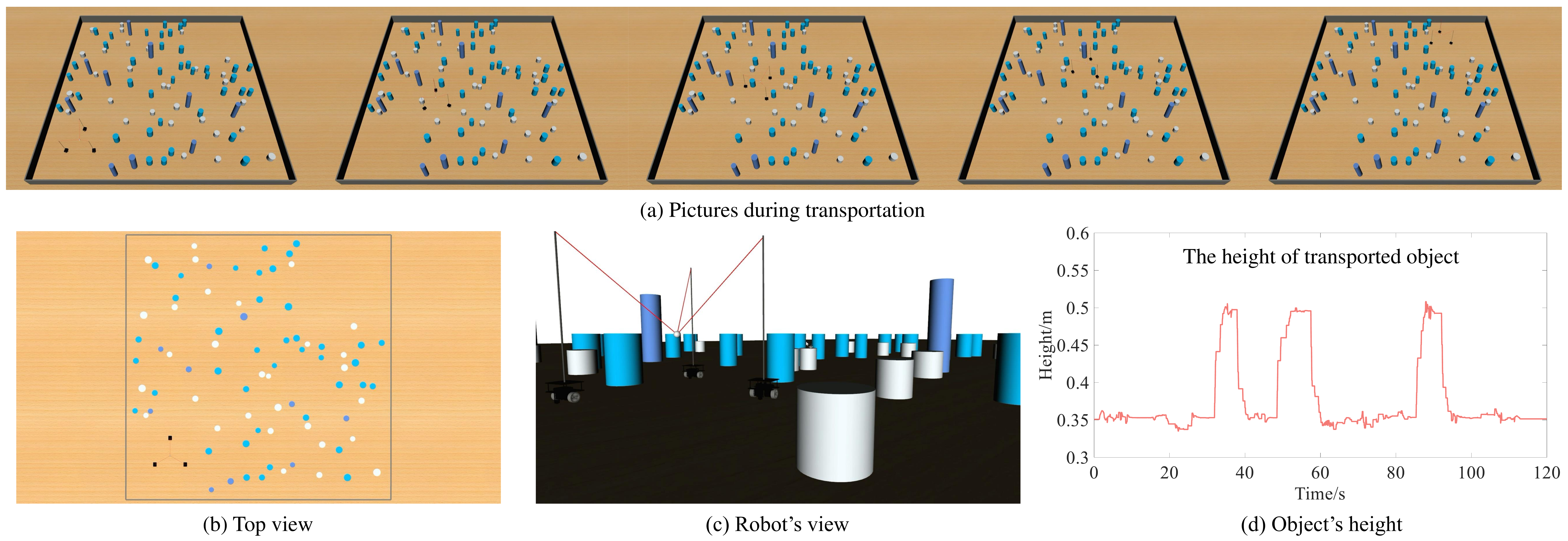}
	\centering
	\caption{
		A multi-mobile robot system composed of three robots transport objects with a deformable sheet in a 10m$\times$10m environment.
		The deformable sheet is simplified to a robots-cables-payload system.
		There are 80 obstacles with random positions and radii. Obstacles of different heights (0.25m, 0.42m, 1m) are distinguished by different colors.
		(a) Pictures during transportation.
		(b) Top view of the whole environment.
		(c) Robot's view. It can be seen that some obstacles can only be bypassed.
		(d) The height of object during transportation.
	}
	\label{fig_3robot}       
\end{figure*}

\begin{figure*}[t]
	\includegraphics[width=\textwidth]{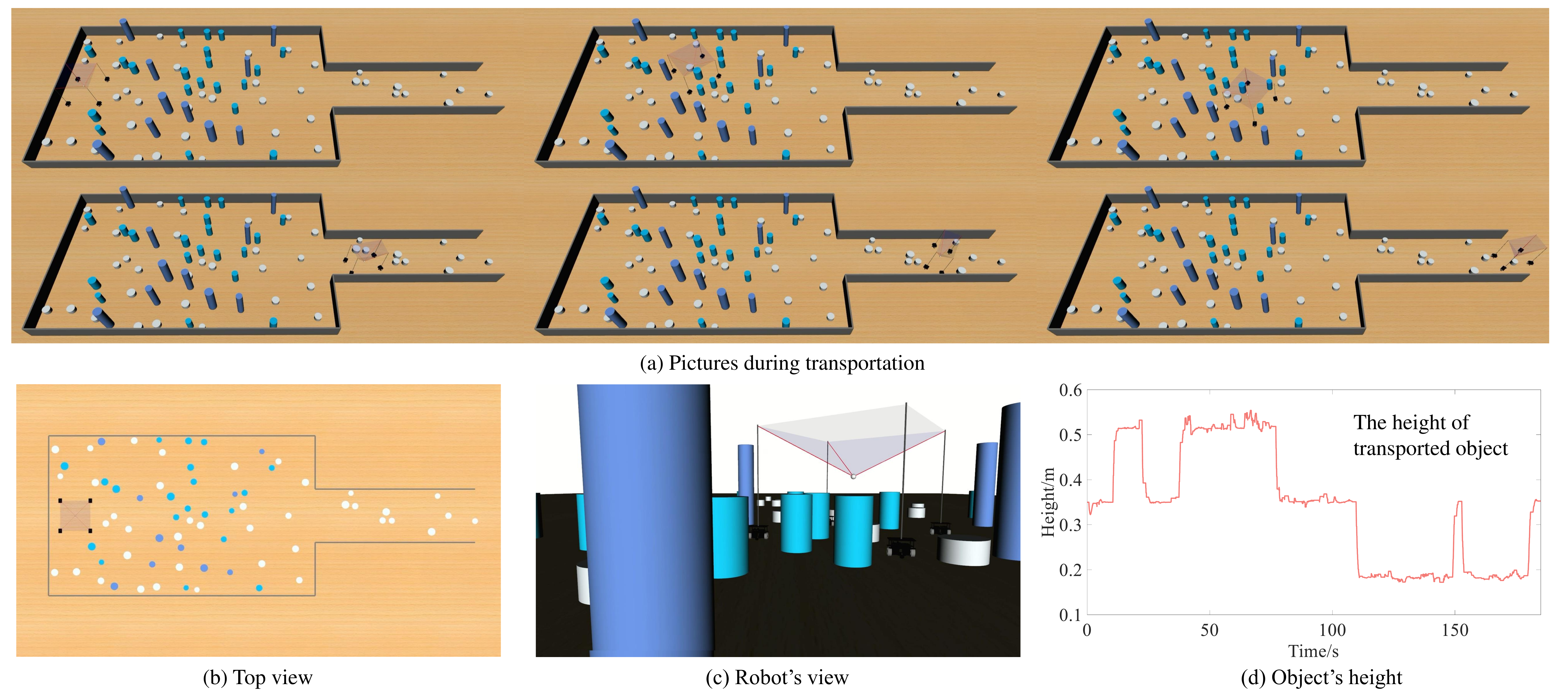}
	\centering
	\caption{
		A multi-mobile robot system composed of four robots transport objects with a deformable sheet in a 16m$\times$6m environment.
		There are 70 obstacles.
		(a) Pictures during transportation.
		(b) Top view of the whole environment.
		(c) Robot's view.
		(d) The height of object during transportation.
	}
	\label{fig_4robot}       
\end{figure*}

\textit{2) Object Transportation without Carrier}

In the second part, mobile robots are required to transport objects without a carrier, which also needs robots to maintain rigid formations.
But different from the first part, there are additional constraints brought by objects.
Robots and environment are the same as Fig.~\ref{fig_rigid}(a).
The transported object is a square with a side length of 1.5m.
There are 30 obstacles randomly generated, and the results are shown in Fig.~\ref{fig_without}.
The system also can reach the goal position without collision, and the average path length of each robot is 18.4439m.
Since obstacles are all higher than the system, constraints of formations and safe transportation can be synthesized, which is that the 2D outline of the whole system does not overlap obstacles. 
Compared with the first part, the obstacle-crossing ability of this system is observably reduced.
It has to merely bypass obstacles and cannot cross them.

\textit{3) Object Transportation with Deformable Sheet}

In the third part, mobile robots are required to transport objects with a deformable sheet, the model of which has been studied in our previous work \cite{hu2022multi}.
Now, formation constraints are no longer rigid, and the system is able to change formations to bypass or cross obstacles depending on the height of transported object.
We first test a three-robot system.
Each side length of the original deformable sheet is 1.6m, and the height of each contact point between robots and the sheet is 1m.
These parameters will be used to calculate the model as described in Eq.\eqref{eq13}.
The environment is also 10m$\times$10m, and the radius of each robot is 0.15m.
The initial position of formation center is (1.5,1.5)m, and the target is (8.5,8.5)m.
There are 80 obstacles randomly generated with random radius between 0.1m and 0.15m, among which 30 are 0.25m in height, 40 are 0.42m in height, and 10 are 1m in height.
Limited by formation constraints and the deformable sheet, the system has to bypass some obstacles.
$\lambda_{s}$ is set as the smallest for the formation of a triangle with each side length of 0.7m.
The results are shown in Fig.~\ref{fig_3robot}, and the deformable sheet is simplified to a robots-cables-payload system as described before.
Obstacles of different heights are distinguished by different colors.
The system can traverse this obstacle-rich environment without collision.
The average path length of each robot is 17.1758m.
In Fig.~\ref{fig_3robot}(c), it can be seen that the system has to change formations to uplift transported object to cross some obstacles.
The height of object during transportation is shown in Fig.~\ref{fig_3robot}(d).

\begin{figure}[t]
	\includegraphics[width=\columnwidth]{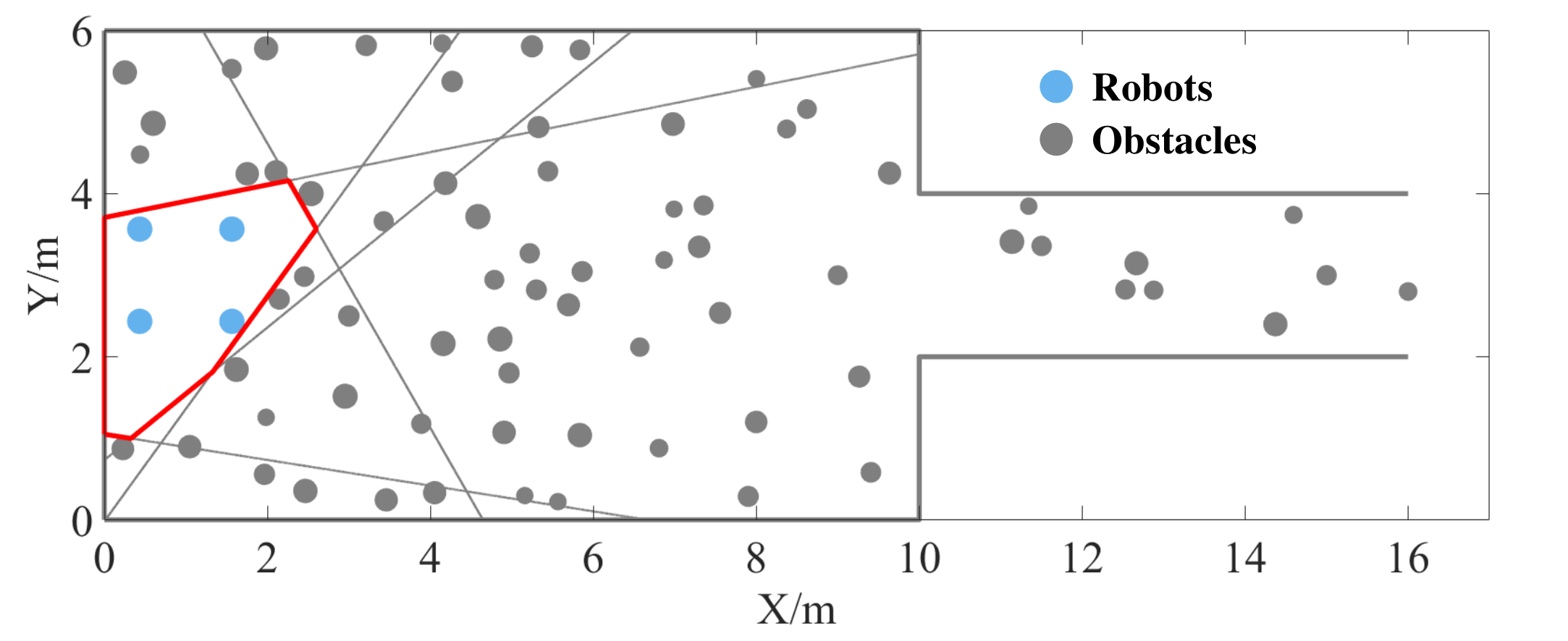}
	\centering
	\caption{
		Motion planning by the method in \cite{alonso2017multi}.  The largest obstacle-free space currently containing the robot system is denoted by the red polygon. Robots can move freely in the red polygon. However, it cannot drive the system to the destination if considering the formation constraints.
	}
	\label{fig_convex}       
\end{figure}

We secondly test a four-robot system in a more complex environment including a narrow corridor.
The environment is 16m$\times$4m and contains 70 obstacles.
$\lambda_{s}$ is set as the smallest for the formation of a square with each side length of 1.1314m.
The results are shown in Fig.~\ref{fig_4robot}.
The average path length of each robot is 29.1217m.
For comparison, we tried the motion planning method in \cite{alonso2017multi}.
The strategy is to find the largest obstacle-free space containing the system, inside which motions of robots are planned.
Due to ignoring heights, the system is considered on the 2D plane.
The result is shown in Fig.~\ref{fig_convex}.
The red polygon is the largest obstacle-free space currently containing the robot system.
Motions of robots can be planned freely in the red polygon which changes with the movement of the system.
It can greatly avoid collisions and maintain desired formations.
However, the method failed after several movements because the obstacle-free space could not be found anymore for the system to move toward the destination.
This method is not appropriate in such obstacle-rich environments.

\subsection{Discussion}

In the planner, we propose a novel mapping between workspace and configuration space of multi-mobile robot systems, which can be efficiently used in planning.
Besides, a boundary densification method is added for facilitating planning.
The planner can deal with different formation constraints and additional constraints brought by special tasks.
We take object transportation as a typical example to test our planner.
Simulation results show that the planner can be applied to obstacle-rich environments.
In the second part, the system is regarded as a whole on the 2D plane to avoid obstacles, which has also appeared in other papers.
Possibly, the problem may be solved by the methods in \cite{ebel2020design,song2021herding,koung2021cooperative,alonso2017multi}.
However, these works have not been tested in such a complex environment.
As for the first and third parts, to the best of our knowledge, no other existing methods can achieve the same effect.
State-of-the-art methods are either unable to maintain formations or cannot be used in these environments.
Despite taking some specific examples in the simulations, the planner can be potentially used in many scenarios as long as formation constraints are determined.
However, our planner also has some limitations.
First, discrete scales are difficult to be chosen in different environments.
There is no existing path in the simulations if the scales are set inappropriately.
The choice of scales dominates our planner.
Moreover, it remains uncertain whether there is a path or scales are inappropriate if planning fails.
Instructively, discrete scales are selected according to control accuracy and obstacle size.
Scales smaller than the control accuracy are not practically meaningful, and only theoretically feasible solutions exist in this case.
Second, the time spent on planning is too heavy, especially when $\mathcal S$ contains many formations.
Now, the planner cannot be applied in an environment with dynamic obstacles.
Since the mapping is full, each valid configuration space of different formations needs to be built.
Indeed, sometimes it is unnecessary and wasteful.

\section{Conclusion and Future Work}\label{sec7}

This paper proposes a novel multi-mobile robot motion planner with excellent obstacle avoidance ability under formation constraints.
The optimal motions of robots can be planned in obstacle-rich environments.
Valid configurations of systems are defined to satisfy formation constraints.
Then, a mapping between workspace and configuration space is introduced with theoretical analysis.
An undirected connected graph is generated to solve the planning problem, on which some existing search methods can be used efficiently.
Moreover, a boundary densification method is used for improving the success rate of planning.
The generality and effectiveness of the planner are proved by implementing different cases.
Despite taking specific examples, the framework of our planner can be potentially employed in other scenarios.

Regarding future work, we will further improve the efficiency of the planner to save time.
Since full mapping is unnecessary sometimes, a hierarchical planning structure may be considered.
And, a trade-off between planning time and approximate optimal solution can also be enhanced.
Obviously, the planner cannot be used in environments with dynamic obstacles.
In this regard, a possible direction is to combine the planner as a global planner with some local obstacle avoidance methods.
When the multi-mobile robot system approaches a dynamic obstacle, local methods will be employed for real-time obstacle avoidance.
More practically, we also will study the disturbance problem when tracking the planned path.
Sensors may be carried by robots in order to react to unforeseen events.
Then, physical experiments can be carried out.

\bibliographystyle{IEEEtran}
\bibliography{References}

\end{document}